\pdfoutput=1

\documentclass[11pt]{article}

\usepackage{ACL2023}

\usepackage{times}
\usepackage{latexsym}

\usepackage[T1]{fontenc}

\usepackage[utf8]{inputenc}

\usepackage{microtype}

\usepackage{inconsolata}
\usepackage{graphicx} 
\usepackage{listings}
\usepackage{spverbatim}
\usepackage{amssymb}
\usepackage{mdframed}
\usepackage{xcolor}
\usepackage{multicol}
\usepackage{multirow}
\usepackage{arydshln}
\usepackage{subcaption}

\title{Rethinking the Instruction Quality: L\resizebox{!}{1.3ex}{IFT} is What You Need}

\author{Yang Xu\textsuperscript{1,2,*} , Yongqiang Yao\textsuperscript{2,*}, Yufan Huang\textsuperscript{2,*}, Mengnan Qi\textsuperscript{2,*}, \\ \bf Maoquan Wang\textsuperscript{2,*}, Bin Gu\textsuperscript{3,4}, Neel Sundaresan\textsuperscript{2}\\
\textsuperscript{1}School of Electronic Information and Electrical Engineering, Shanghai Jiao Tong University \\
\textsuperscript{2}Microsoft Cloud and AI \\
\textsuperscript{3}School of Artificial Intelligence, Jilin University \\
\textsuperscript{4}Mohamed bin Zayed University of Artificial Intelligence\\
\tt xuyang2018@sjtu.edu.cn \\
\tt \{yongqiangyao, yufanhuang, mengnanqi, maoquanwang\}@microsoft.com \\
}

\begin{document}
\maketitle
\begin{abstract}
Instruction tuning, a specialized technique to enhance large language model (LLM) performance via instruction datasets, relies heavily on the quality of employed data. Existing quality improvement methods alter instruction data through dataset expansion or curation. However, the expansion method risks data redundancy, potentially compromising LLM performance, while the curation approach confines the LLM's potential to the original dataset. Our aim is to surpass the original data quality without encountering these shortcomings. To achieve this, we propose  \textit{LIFT} (\textbf{L}LM \textbf{I}nstruction \textbf{F}usion \textbf{T}ransfer), a novel and versatile paradigm designed to elevate the instruction quality to new heights. LIFT strategically broadens data distribution to encompass more high-quality subspaces and eliminates redundancy, concentrating on high-quality segments across overall data subspaces. Experimental results demonstrate that, even with a limited quantity of high-quality instruction data selected by our paradigm, LLMs not only consistently uphold robust performance across various tasks but also surpass some state-of-the-art results, highlighting the significant improvement in instruction quality achieved by our paradigm.

\end{abstract}

\section{Introduction}

In recent years, Large Language Models (LLMs) have gained prominence for their remarkable effectiveness in natural language comprehension tasks \citep{openai-2023-gpt4, yang-2023-harnessing}. High-quality pretrained LLMs are readily available, facilitating their customization for versatile applications. One popular fine-tuning approach, known as instruction tuning \citep{wei-2022-finetuned, ouyang-2022-training}, involves fine-tuning pre-trained LLMs using datasets accompanied by natural language instructions. Its relative simplicity and affordability make it a preferred method for improving LLMs' performance on specific tasks.

\begin{figure}[ht]
    \centering
    \includegraphics[width=\linewidth]{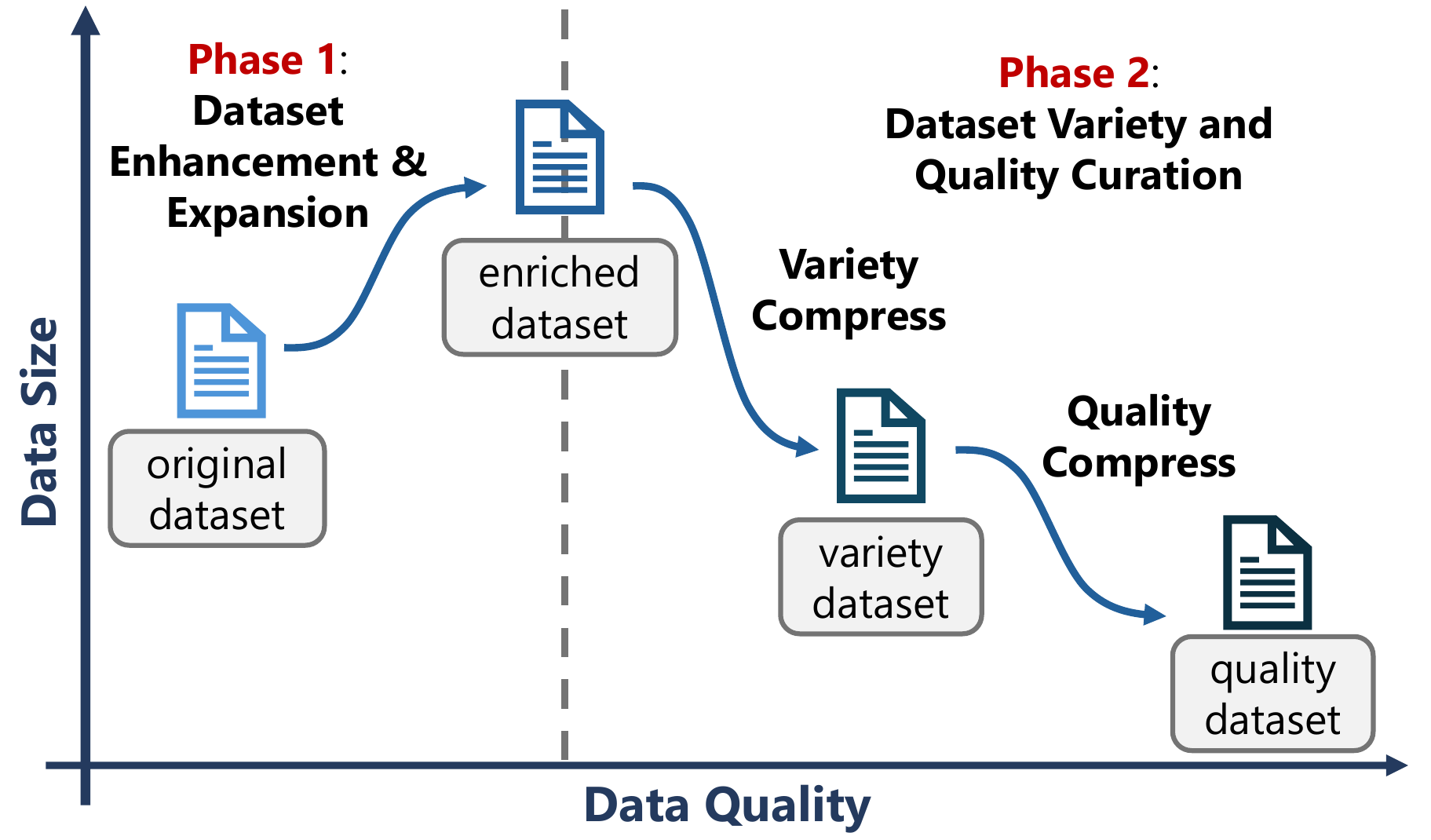}
    \caption{Instruction Dataset Curation Paradigm \textbf{LIFT}}
    \label{fig:paradigm}
\end{figure}

The effectiveness of instruction tuning heavily relies on the quality of the instruction dataset. Currently, two prevalent methods for constructing instruction datasets are manual crafting \citep{ouyang-2022-training} and LLM-generated \citep{chiang-2023-vicuna, rohan-2023-alpaca}. However, these methods often fall short in terms of quality. Human-crafted datasets depend on human annotators to generate a substantial corpus with human instructions, leading to the instruction dataset lacking detailed context and explanation. It may also contain vague or objective descriptions. On the other hand, LLM-generated methods use advanced LLMs to generate or complete instructions and responses but lack supervision regarding the diversity and quality of the generated data.
 
The concern regarding instruction datasets' quality has prompted researchers to explore methods for improving their quality. Current approaches to quality enhancement can be broadly categorized into two groups: data expansion and data curation. Data expansion methods involve leveraging advanced LLMs, such as GPT-4, with a suitable prompt template to generate new instructions and corresponding answers based on the original dataset \citep{xu-2023-wizardlm, luo-2023-wizardcoder, rohan-2023-alpaca}. On the other hand, data curation methods involve the meticulous selection of high-quality data from the original dataset based on specific quality evaluation criteria \citep{zhou-2023-lima, du-2023-mods}. We posit that both methods essentially function as data distribution transfers: expansion enables the distribution to cover a broader range of data subspaces, typically characterized by higher quality, while curation concentrates the distribution on a higher-quality subset of the original dataset.

However, both current methods have limitations that impede their capacity to further enhance performance. Expansion methods introduce redundancy into the dataset as the newly generated instructions are typically based on the original ones. The effectiveness of curation methods heavily relies on the quality of the original dataset, limiting the quality of the curated dataset. These limitations force these methods to rely on specific expansion or curation strategies to achieve better performance on certain benchmarks, at the expense of losing the ability to generalize the approach.

In this paper, we delve deeper into the quality distribution transfer process to address the mentioned problems. We propose a novel paradigm for improving LLM instruction quality, termed \textbf{LIFT} (\textbf{L}LM \textbf{I}nstruction \textbf{F}usion \textbf{T}ransfer). LIFT is designed to amalgamate the advantages of data expansion and curation, mitigating their shortcomings to generate a diverse and high-quality dataset while significantly reducing quantity. Our paradigm consists of two phases. Firstly, we employ "Dataset Distribution Expansion", broadening the data distribution to cover more high-quality subspaces. Then, we utilize "Dataset Variety and Quality Curation" to eliminate redundancy and densify the data distribution, focusing on the high-quality segments of overall data subspaces. To validate the effectiveness of LIFT, we employ the finally curated instructions for fine-tuning open-source LLMs. Through extensive experiments evaluating the performance of these fine-tuned LLMs in both natural language understanding (NLU) tasks and code generation tasks, the results consistently demonstrate that LLMs maintain robust performance even with a limited quantity of high-quality instruction data. In certain cases, they even outperform models trained on larger datasets. To summarize, our main contribution are:

\begin{itemize}

\item We propose a highly effective and versatile paradigm, LIFT, which challenges the conventional single-mode enhancement for instruction datasets. LIFT rethinks data quality by focusing on data distribution transfer. It aims to elevate the quality of the instruction dataset to new heights, overcoming redundancy and quality limitations present in current methods.

\item Throughout the expansion and curation phases of the paradigm, we prioritize both variety and quality as essential goals for enhancement. Unlike existing works that concentrate on only one stage, we posit that these characteristics play a crucial role in bringing more high-quality data into the final curated dataset.

\item Our extensive experimental results demonstrate that, with a significantly reduced quantity of high-quality instruction data selected by our paradigm, LLMs consistently achieve nearly-SOTA or SOTA performance across various benchmarks. This provides valuable insights, suggesting that a selective approach based on the principles of data distribution transfer is not only more effective but also cost-effective compared to the indiscriminate feeding of large volumes of data into LLMs during training.

\end{itemize}

\section{Related Works}

\subsection{Instruction Dataset Expansion}



\subsection{Instruction Dataset Curation}


To enhance the model's performance following instruction fine-tuning, some researchers have concentrated on filtering out low-quality data during the fine-tuning stage. LIMA \citep{zhou-2023-lima} demonstrates that fine-tuning a robust pre-trained language model on 1000 high-quality, human-curated examples can yield remarkable and competitive results. Instruction Mining \citep{cao-2023-instruction} introduces a linear rule for selecting high-quality instruction data, eliminating the need for human annotation, as seen in LIMA. \citet{du-2023-mods} present a model-oriented data selection (MoDS) approach, which selects instruction data based on a new criteria considering three aspects: quality, coverage and necessity. \citet{li-2023-fromqt} introduce a self-guided methodology for LLMs to autonomously discern and select cherry samples from vast open-source datasets, effectively minimizing manual curation and potential cost.

In contrast to the aforementioned works, which focused solely on a single LLM task, our paradigm offers a versatile solution for effectively selecting diverse and high-quality data from the raw dataset used for fine-tuning, encompassing both NLU tasks and code-related tasks.

\section{LLM Instruction Fusion Transfer}


\subsection{Data Distribution Transfer}

Current methods for enhancing instruction quality, whether through data expansion or curation, do enhance the original dataset to some extent, thereby improving the performance of LLMs on specific tasks when using these enhanced datasets for instruction tuning. However, the effectiveness of these methods is constrained by inherent limitations. To scrutinize these limitations and explore innovative approaches to break from conventional enhancement modes, we propose a novel perspective for rethinking instruction data quality: \textbf{data distribution transfer}.

\subsubsection{Analysis of Existing Methods}

Our hypothesis is that, during the process of enhancing instruction quality, there is a transfer of data distribution from the original dataset to the final enhanced dataset. This transfer increases the quantity or proportion of high-quality data. In data expansion methods, generating high-quality instructions based on the original ones effectively extends the coverage of the high-quality data subspace within the original data distribution, thereby increasing the quantity of high-quality data in the final distribution. On the other hand, in data curation methods, by using carefully designed quality evaluation metrics, low-quality components are removed from the final distribution, directing the distribution to concentrate on high-quality data and increasing its proportion in the final distribution. 

From this perspective, we can easily identify the limitations in these methods. The areas around the original instructions may contain similar ones, leading to redundancy in the final distribution. Moreover, low-quality instructions and those derived from them persist in the final distribution, maintaining a proportion similar to the original dataset. On the other hand, curation methods select a portion of high-quality instructions from the original dataset, resulting in a decrease in the total number of high-quality instructions. If the original dataset has a limited number of high-quality instructions, the quality of the curated dataset will significantly decrease.

\subsubsection{Fusing Expansion and Curation}

Analyzing the data distribution transfer patterns of expansion and curation, we suggest that their integration can effectively address their individual limitations. Data expansion broadens subspaces, enabling the curation method to explore beyond the original distribution. Conversely, data curation assists in identifying duplicates and low-quality items from the expansion, contributing to a more concentrated and refined distribution. 

Building on these insights, we introduce a novel paradigm called LIFT (LLM Instruction Fusion Transfer). Comprising two phases, this paradigm orchestrates the distribution of instruction data as follows: in the "Dataset Distribution Expansion" phase, we broaden the data distribution to encompass more diverse and high-quality subspaces, acknowledging the presence of duplications at this stage. Subsequently, in the "Dataset Variety and Quality Curation" phase, we systematically eliminate redundancy and low-quality elements, creating a densified distribution for the final curated dataset. These two phases are intricately connected, ensuring a smooth transfer of data from the original dataset to the final curated dataset.

\subsection{Paradigm LIFT}

As mentioned in the above section, our paradigm LIFT follows a two-stage structure. In both stages, we value the diversity and quality as the crucial criterion since we believe the "Dataset Distribution Expansion" and "Dataset Variety and Quality Curation" equally contribute to the quality enhancement.

\subsubsection{Dataset Distribution Expansion}

The goal of dataset distribution expansion is to encompass a more diverse and high-quality range of data within the distribution, while ensuring a certain distance from the original instructions. To achieve this, it is crucial to employ carefully designed instruction-generation prompts. Drawing inspiration from the instruction rewrite method proposed by \citeauthor{xu-2023-wizardlm}, our approach focuses on generating diverse and intricate instructions. We guide GPT-4 to act as a prompt rewriter, generating challenging instructions based on specified generation rules. Considering the variation in content for NLU and code generation tasks within the instruction dataset, we configure distinct settings for GPT prompts to enhance complexity. For further details, refer to Appendix \ref{sec:prompt-template}. We iterate this process for $k$ rounds, merging the expanded datasets with the original dataset to create the final expanded dataset.

\subsubsection{Dataset Variety and Quality Curation}

\begin{figure*}[ht]
    \centering
    \includegraphics[width=0.95\linewidth]{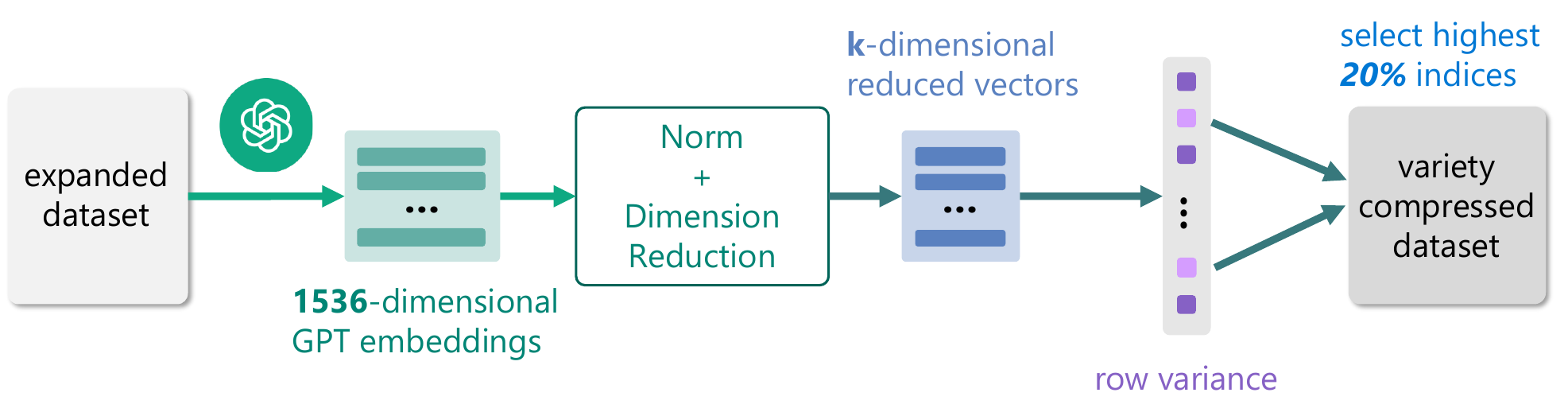}
    \caption{Variety Compress with Dimension Reduction and Row Variances}
    \label{fig:variety}
\end{figure*}

\begin{figure*}[ht]
    \centering
    \includegraphics[width=0.95\linewidth]{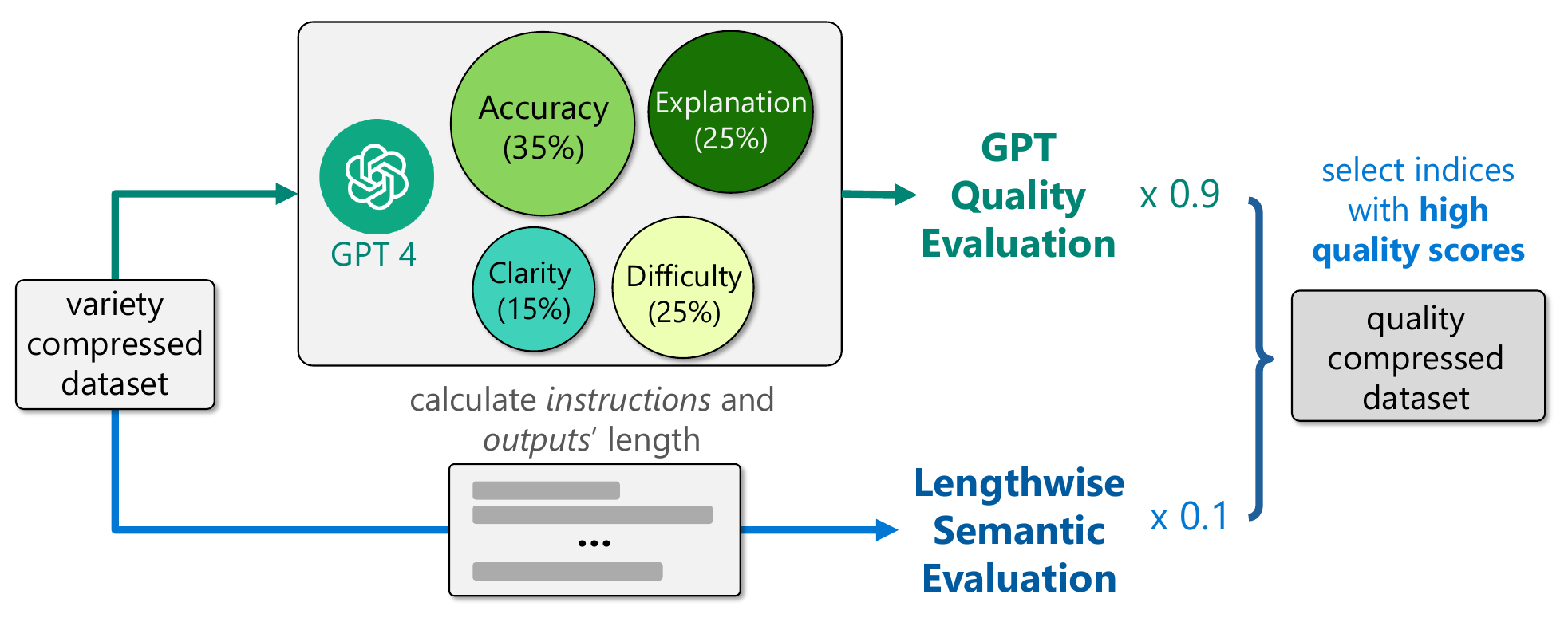}
    \caption{Quality Compress with GPT Quality Evaluation and Lengthwise Semantic Evaluation}
    \label{fig:quality}
\end{figure*}

An effective curation method ought to eliminate duplicated or low-quality instructions from the original dataset, while preserving representative and high-quality ones. To meet this criterion, curation should be approached with meticulous attention to both variety and quality.

Current variety curation typically involves clustering methods such as k-means or spectral clustering, which initially segment the original data distribution into several small groups, followed by the selection of representative items from these groups. We argue that this approach may lack generalizability and be less effective when dealing with new datasets. This is because these methods require prior knowledge of the number of clusters, and choosing cluster numbers that are either too large or too small may reduce their effectiveness in selecting representatives.

Our variety curation method take another route, as depicted in Fig. \ref{fig:variety}. Initially, GPT generates embeddings with 1536 dimensions for each item, which proves unwieldy for analysis. To address this, we aim to reduce the embedding dimension and devise a method to represent data differentiation. We achieve this by calculating the covariance matrix of the given features and performing eigenvalue decomposition on the covariance matrix to obtain eigenvalues and eigenvectors. We then choose the top $k$ eigenvectors corresponding to the largest $k$ eigenvalues, where $k$ is the target reduced dimension. This process allows us to analyze row variance on the dimension-reduced features to identify items with significant differences. Row variance measures variability in the reduced space, and high variance suggests substantial positional changes along that dimension, aiding in identifying diverse data points. We select items with the highest 20\% row variances to construct the variety-curated dataset. This method doesn't require any prior statistical knowledge of the dataset, making it versatile and effective for all tasks.

Following variety curation is the quality curation phase, where we discern high-quality instruction data. Rating the quality of instruction data is challenging due to the lack of official quantitative metrics. Given the dataset's size and associated costs, employing professional annotators for individual scoring is impractical. Therefore, we turn to GPT-4 as an instruction scorer, generating \textbf{GPT quality scores} for each instruction. GPT-4 evaluates the instruction data across four dimensions: accuracy, explanation, clarity, and difficulty. The proportions assigned to these metrics are based on their contributions to overall instruction quality. The template guiding the GPT-4 scores is presented in Appx. \ref{sec:score-template}.

In our practical experience, we observed that GPT-4 consistently assigns high total scores to all instruction data, even when explicitly instructed not to do so, posing a challenge by reducing differentiation in GPT quality scores. To address this, we implement the following steps for more reasonable and differentiated scores:

\textbf{Require scores along with detailed explanations}. Instructing GPT-4 to provide a comprehensive rationale along with a score yields more reasonable results, preventing indiscriminate high-level assessments. While we may not use these explicit explanations, mandating GPT-4 to articulate its reasoning offers an additional self-checking opportunity.

\textbf{Provide few-shots scoring examples}. Before scoring instruction items, we present manually scored examples as guidelines. Offering three examples with scores representing poor, average, and high quality helps GPT-4 recognize low-quality data and understand how to appropriately score it.

Furthermore, recognizing that the total length of instruction data indicates information richness, we incorporate this into the quality score calculation, too. Unlike other metrics, length is quantitatively analyzable. A positively correlated mapping function derives a \textbf{lengthwise semantic score} based on the instruction data's length. Combining the GPT quality score and lengthwise semantic score produces the final quality score. Instruction items with high-quality scores compose the final quality-compressed dataset, as illustrated in Fig. \ref{fig:quality}. In Appx. \ref{sec:quality-distribution}, we present an overview of the total quality score distributions across both tasks. Our meticulous efforts ensure substantial differentiation in quality scores, allowing precise identification and selection of high-quality data.

\section{Experiments}

To validate the effectiveness of our paradigm, we apply our method to two extensively studied tasks: Code Generation tasks and NLU tasks. Subsequently, we conduct comprehensive experiments to evaluate the performance of our approach.

\subsection{Experiments Setup}

\subsubsection{Basic Foundation Models and Base Datasets}

For code generation tasks, we leverage StarCoder 15B \citep{li-2023-starcoder}, a widely-used large code language model trained on diverse sources, including 80+ programming languages, Git commits, GitHub issues, and Jupyter notebooks. Our base dataset, Code Alpaca \citep{sahil-2023-codealpaca}, consists of 20k instruction-following code instances for fine-tuning Code LLMs.

In NLU tasks, we employ the state-of-the-art model Mistral 7B \citep{jiang-2023-mistral}, known for superior performance compared to other 7B models and outperforming larger models in certain benchmarks. Our base dataset for NLU tasks is the Open Platypus dataset \citep{lee-2023-platypus}, comprising 25k curated examples focused on enhancing LLMs' STEM and logic knowledge. This dataset, composed of 11 open-source datasets, mainly contains human-designed questions, with only 10\% generated by an LLM.

The instruction tuning implementation details for both tasks can be found in Appx. \ref{sec:implementation_details}

\subsection{Benchmarks and Metrics}

We have selected six widely-used benchmarks across the two tasks. For code generation tasks, we incorporate HumanEval and MBPP. And for NLU tasks, our selected benchmarks include HellaSwag, ARC Challenge, TruthfulQA, and MMLU. The details of these selected benchmarks can be found in Appx. \ref{sec:benchmarks}. 

In code generation tasks, our metric of choice is pass@k, defined in the same manner as by \citeauthor{chen-2021-evaluating}. The formula for calculating pass@k is presented as:

\[pass@k := \mathbb{E}_{problems}[1-\frac{C(n-c,k)}{C(n,k)}]\]

Here, $n$ represents the number of generated answers for each question, and $c$ denotes the number of correct answers for each question. In our experiments, we choose pass@1 as the metric.

For NLU tasks, we opt for accuracy as the metric, 
aligning with the approach adopted by other researchers.

\subsection{Experiment Results}

We compare models fine-tuned on our paradigm LIFT's final curated dataset with other state-of-the-art (SOTA) pretrained LLMs and instruction-tuned LLMs across both tasks. The details of the selected models for comparison is presented in Appx. \ref{sec:compared llms}

\subsubsection{Code Generation Tasks}

As depicted in Tab. \ref{tab:compared_code}, our paradigm's fine-tuned model outperforms most of the models in code generation tasks. While our model is nearly 2\% lower than the current state-of-the-art 15B model, WizardCoder, on both benchmarks, it's important to note that our paradigm utilizes only about one-eighth of the instruction data used by WizardCoder. Considering the size of the instruction dataset, our paradigm demonstrates robust performance, showcasing the ability to achieve performance close to the state-of-the-art with a significantly smaller amount of data.

\begin{table}[ht]
    \caption{LLMs Performance Comparison in Code Generation Tasks}
    \centering
    \resizebox{\linewidth}{!}{
        \begin{tabular}{lccc}
            \hline
            \textbf{Model}& \textbf{Data Size} & \textbf{HumanEval} & \textbf{MBPP}\\
            \hline
            CodeT5+ & \multirow{3}{*}{-\textsuperscript{*}} & 0.309 & - \\
            CodeLLaMA &  & 0.360 & 0.470 \\
            StarCoder &  & 0.336 & 0.436 \\
            \hdashline
            InstructCodeT5+ & 20k & 0.350 & - \\
            WizardCoder & 78k & \textbf{0.573} & \textbf{0.518} \\
            \hdashline
            Random & 10k & 0.381 & 0.431 \\
            \textbf{LIFT} & \textbf{10k} & 0.550 & 0.495 \\
            \hline
            \multicolumn{4}{l}{* \textit{Pretrained models}}
            
        \end{tabular}
    }
    \label{tab:compared_code}
\end{table}

\subsubsection{Natural Language Understanding Tasks}

\begin{table*}[ht]
    \caption{LLMs Performance Comparison in NLU Tasks}
    \centering
    \begin{tabular}{lccccc}
        \hline
        \textbf{Model} & \textbf{Fine-tuning Data Size} & \textbf{HellaSwag} & \textbf{ARC} & \textbf{TruthfulQA} & \textbf{MMLU} \\
        \hline
        LLaMA-7B & \multirow{5}{*}{\textit{Pretrained}} & 0.778 & 0.509 & 0.343 & 0.357\\
        LLaMA-13B & & 0.809 & 0.561 & 0.395 & 0.476 \\
        LLaMA2-7B & & 0.771 & 0.432 & 0.333 & 0.444 \\
        LLaMA2-13B & & 0.807 & 0.488 & 0.419 & 0.556 \\
        Mistral-7B & & 0.823 & 0.602 & 0.426 & 0.627 \\
        \hdashline
        Vicuna-7B & 70k conversations & 0.775 & 0.537 & 0.489 & 0.456\\
        Vicuna-13B & 70k conversations & 0.801 & 0.530 & 0.518 & 0.513 \\
        WizardLM-7B & 70k instructions & 0.771 & 0.516 & 0.447 & 0.427 \\
        WizardLM-13B & 70k instructions & 0.777 & 0.572 & 0.505 & 0.523 \\
        Platypus2-13B & 25k instructions& 0.826 & 0.613 & 0.449 & 0.567 \\
        Camel-Platypus2-13B & 25k instructions & 0.836 & 0.608 & 0.496 & 0.565 \\
        Stable-Platypus2-13B & 25k instructions & 0.822 & 0.627 & \textbf{0.525} & 0.583 \\
        \hdashline
        Random Selection-7B & 15k instructions & 0.820 & 0.607 & 0.438 & 0.625 \\
        \textbf{LIFT-7B} & \textbf{15k} instructions & \textbf{0.844} & \textbf{0.643} & 0.490 & \textbf{0.645} \\
        \hline
    \end{tabular}
    \label{tab:compared_nlu}
\end{table*}

Tab. \ref{tab:compared_nlu} presents the NLU task comparison results. Notably, our final instruction-tuned model consistently outperforms other 7B models on all benchmarks. Even among 13B models, our model outperforms in all benchmarks except TruthfulQA. With only 7 billion parameters and 15k instructions, significantly fewer than other instruction-tuned models, our model achieves the highest average benchmark score at 0.656.

We also compared our paradigm's final curated dataset with a randomly selected dataset of the same size in both tasks. The results demonstrate that merely reducing the dataset quantity, without accounting for the diversity and quality of the data, does not lead to performance improvement. These experiments affirm our paradigm's versatile effectiveness in code generation and NLU tasks. The paradigm excels in generating diverse, high-quality data, leveraging it in the instruction-tuning process to achieve SOTA or near-SOTA performance.

\subsection{Paradigm Ablation Experiments Results}

\label{sec:paradigm_experiments}

Our paradigm ablation experiment begins with the original base dataset serving as the input for LIFT. Subsequently, we generate the expanded dataset, variety-compressed dataset, and the quality-compressed dataset. These datasets are then utilized for fine-tuning the basic foundation models. We assess the benchmark performance of these models to validate the effectiveness of each component of our paradigm.

\subsubsection{Code Generation Tasks}

Tab. \ref{tab:paradigm_code} provides an overview of the paradigm experiments conducted on code generation tasks. For data expansion, we repeatedly perform the expansion step two times, resulting in a 60k size instruction dataset. 

\begin{table}[ht]
    \caption{Paradigm Ablation Experiment Results (Pass@1) in Code Generation Tasks}
    \centering
    \resizebox{\linewidth}{!}{
        \begin{tabular}{lccc}
            \hline
            \textbf{Dataset Type}& \textbf{Size} & \textbf{HumanEval} & \textbf{MBPP}\\
            \hline
            Base Dataset & 20k & 0.4091 & 0.4662 \\
            Data Expansion & 60k & 0.5342 & 0.4874 \\
            Variety Compress & 12k & 0.5412 & 0.4887 \\
            Quality Compress & \textbf{10k} & \textbf{0.5503} & \textbf{0.4949} \\
            \hline
        \end{tabular}
    }
    \label{tab:paradigm_code}
\end{table}

The table illustrates our paradigm leads to a significant enhancement in the performance of the fine-tuned model across both benchmarks. Notably, our final compressed dataset, although roughly half the size of the original dataset, outperforms the latter by nearly 15\% on the HumanEval and 3\% on the MBPP. Furthermore, we observe a consistent improvement in model performance on both benchmarks after each step of the paradigm. This implies that the instruction's quality is steadily increasing at each stage. 
These results strongly affirm that, compared to the quantity of the instruction dataset, the variety and quality of the dataset play a more significant role in enhancing performance.

\subsubsection{Natural Language Understanding Tasks}

\begin{table*}[ht]
    \caption{Paradigm Ablation Experiment Results in NLU Tasks}
    \centering
    \begin{tabular}{lccccc}
        \hline
        \textbf{Dataset Type} & \textbf{Dataset Size} & \textbf{HellaSwag} & \textbf{ARC Challenge} & \textbf{TruthfulQA} & \textbf{MMLU} \\
        \hline
        Base Platypus Dataset & 25k & 0.82788 & 0.61543 & 0.44481 & 0.62619\\
        Data Expansion & 100k & 0.83308 & 0.62372 & 0.44718 & 0.63065 \\
        Variety Compress & 20k & 0.83947 & 0.63311 & 0.45615 & 0.64199 \\
        Quality Compress & \textbf{15k} & \textbf{0.84415} & \textbf{0.64334} & \textbf{0.48985} & \textbf{0.64519} \\
        \hline
    \end{tabular}
    \label{tab:paradigm_nlu}
\end{table*}

Tab. \ref{tab:paradigm_nlu} presents our paradigm experiment results on four NLU benchmarks. For data expansion, we iteratively perform the expansion step three times, resulting in a 100k size instruction dataset. 

The table results affirm our paradigm's effectiveness in NLU. Despite a reduction in size by 10k instances compared to the original dataset, our final compressed dataset maintains robust performance, showing improvements ranging from nearly 2\% to 4\% on each benchmark. The observed performance improvement pattern in code generation tasks also applies to NLU tasks, with each step enhancing data quality.

It's crucial to note that unlike the Code Alpaca \citep{sahil-2023-codealpaca} dataset, the Open Platypus \citep{lee-2023-platypus} dataset for NLU tasks is already carefully curated.
The results for this dataset underscore that our paradigm is effective not only for LLM-generated datasets but also in elevating the quality of already high-quality datasets, contributing to improved fine-tuned model performance while reducing the dataset size.

\section{Discussions}

\subsection{Composition of The Final Curated Dataset}

\begin{figure}[ht]
  \centering
  \begin{subfigure}[b]{\linewidth}
    \centering
    \includegraphics[width=0.75\textwidth]{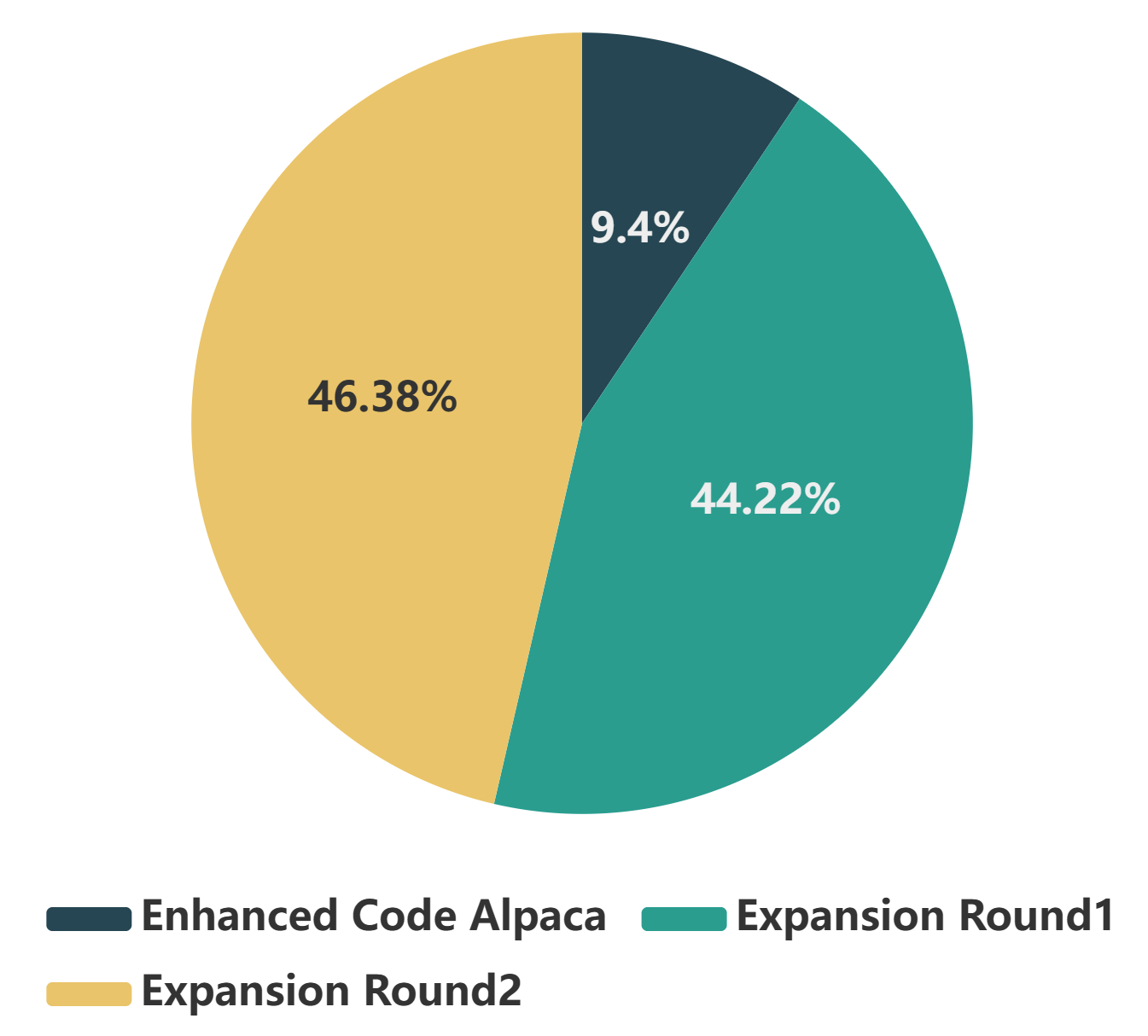}
    \caption{Source of the final curated 10k dataset in code generation tasks}
    \label{fig:composition_10k}
  \end{subfigure}
  \hfill
  \begin{subfigure}[b]{\linewidth}
    \centering
    \includegraphics[width=0.75\textwidth]{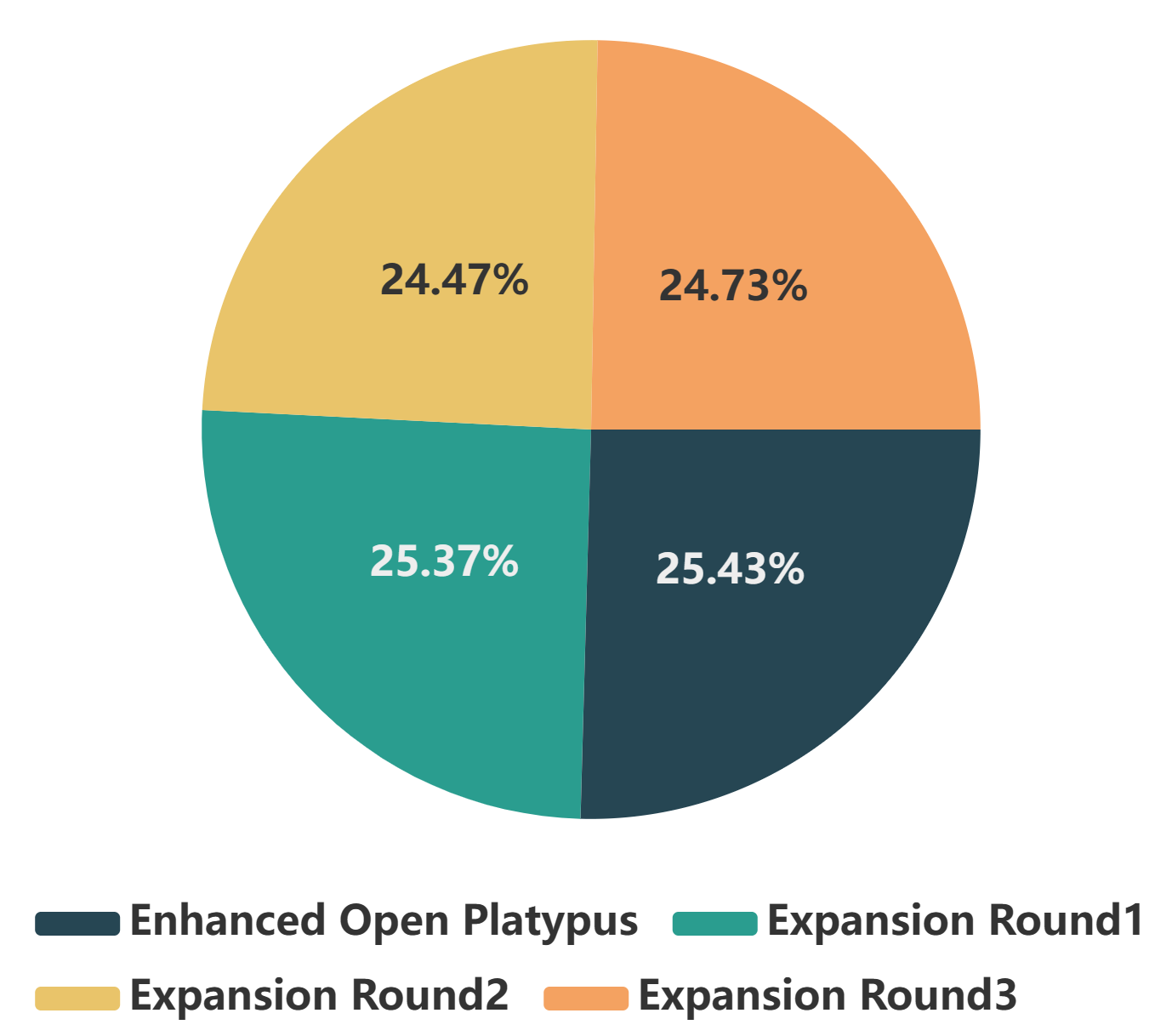}
    \caption{Source of the final curated 15k dataset in NLU tasks}
    \label{fig:composition_15k}
  \end{subfigure}
  \caption{Composition of The Final Curated Dataset}
  \label{fig:composition}
\end{figure}

We take a step further to analyze the composition of the final curated dataset, unraveling the origins of diverse and high-quality instruction items. Fig. \ref{fig:composition} presents the source proportions of the final curated dataset for code generation and NLU tasks, yielding several noteworthy conclusions.

For LLM-generated instruction datasets like Code Alpaca, only a small proportion of the final dataset emanates from the original dataset (Fig. \ref{fig:composition_10k}). The majority of high-quality data is derived from our paradigm's first step—the expanded dataset. This emphasizes our paradigm's significant role in generating a diverse and high-quality dataset, especially for datasets without meticulous curation.

In contrast, for a curated and high-quality instruction dataset like Open Platypus, the portion of the original dataset in the final dataset increases (Fig. \ref{fig:composition_15k}). The proportions of the final curated dataset in NLU tasks reveal an almost equal distribution among the four sub-datasets, demonstrating that even for an initially high-quality dataset, our paradigm excels in generating and selecting numerous high-quality data points based on the original dataset.

These conclusions affirm that the bulk of the final dataset primarily comprises data from the expanded dataset. While proportions of original dataset data contributing to the final dataset may vary based on the original dataset's quality, our paradigm consistently showcases its ability to extract high-quality segments from the original dataset and augment them with diverse and high-quality instructions.

\subsection{GPU Hours and Carbon Emission}

By compressing the size of the instruction dataset, we aim to reduce the GPU hours required for instruction tuning, contributing to a subsequent decrease in carbon emissions. Tab. \ref{tab:gpu_hours} illustrates the impact of different dataset sizes on GPU hours and CO\textsubscript{2} emissions.
GPU hours are calculated under the same settings of training epoch and batch size, while carbon emissions are computed using an online machine learning CO\textsubscript{2} calculator\footnote{\url{https://mlco2.github.io/impact/\#co2eq}}.

\begin{table}
    \caption{Analysis of GPU hours and Carbon Emission with different dataset size. GPU hours in the table are measured in \textit{hour}, CO\textsubscript{2} emission is in \textit{kg CO\textsubscript{2} eq.}}
    \centering
    \begin{tabular}{lcc}
        \hline
        \textbf{Dataset Size}& \textbf{GPU Hours} & \textbf{CO\textsubscript{2} Emission}\\
        \hline
        \multicolumn{3}{l}{\textbf{\textit{Code Generation Tasks}}}\\
        \hline
        Original 20k & 50.82 & 4.58 \\
        Expanded 60k & 185.6 & 16.7 \\
        Curated 10k & 31.6 & 2.84 \\
        \hline
        \multicolumn{3}{l}{\textbf{\textit{NLU Tasks}}}\\
        \hline
        Original 25k & 40.24 & 3.62 \\
        Expanded 100k & 149.76 & 13.48 \\
        Curated 15k & 23.71 & 2.13 \\
        \hline
    \end{tabular}
    \label{tab:gpu_hours}
\end{table}

The table shows a substantial reduction in GPU hours and lower carbon emissions when fine-tuning with the final curated dataset. 
This comparison demonstrates that our paradigm not only accelerates fine-tuning but also promotes environmental sustainability while maintaining robust high performance.

\subsection{Limitations and Future Works}

Our paradigm has limitations requiring further research for resolution.


\textbf{Redundancy in Data Expansion}: The first step involves data expansion, introducing some data redundancy as the new instructions are primarily based on rewriting the original ones. Despite mitigation in the Variety Compress step, the impact of duplicated instructions from the expansion phase should be considered. We aim to explore strategies for expanding data that are less reliant on the original dataset, implementing stringent data supervision strategies for generating diverse instructions.

\textbf{Subjectivity in Quality Evaluation}: Quality evaluation relies heavily on the GPT Quality Evaluation, and despite careful criteria, additional statistical analysis beyond the length factor could enhance the precision of high-quality data selection. Future work will involve enhancing quality evaluation metrics beyond the current focus on instruction length.







\section{Conclusions}

This paper introduces a novel paradigm \textbf{LIFT} for curating large language model instruction datasets to enhance the quality and diversity of data for fine-tuning LLMs. The paradigm unfolds in three steps: Data Enhancement and Expansion, Variety Compress, and Quality Compress.

Experimental results shows that our fine-tuned models consistently achieve either SOTA or nearly SOTA performance across both code generation and natural language understanding tasks. These experiments affirm the paradigm's versatile effectiveness, demonstrating its ability to generate diverse and high-quality data. Integrating this curated data into the instruction-tuning process empowers LLMs to achieve superior performance across various benchmarks.


\bibliography{anthology,main}
\bibliographystyle{acl_natbib}

\appendix

\section{GPT Prompt Templates For Data Expansion}

\label{sec:prompt-template}

We adopt two different prompt settings for natural language understanding tasks and code generation tasks. 

\subsection{Natural Language Understanding Tasks}

\begin{spverbatim}
SYSTEM MESSAGE:
I want you act as a professional prompt re-writer. 
USER PROMPTS:
Your objective is to rewrite a given prompt into a more complex version using data format to make those famous AI systems more difficult to handle. But the rewritten prompt must be reasonable and must be understood and responded by humans. 
You can increase the difficulty using, but not limited to, the following methods:
(1) The depth and breadth of the inquiry can be increased.
(2) Replace general concepts with more specific concepts.
(3) If original problem can be solved with just a few simple thinking processes, you can rewrite it to explicitly request multiple-step reasoning.

#Instruction#
{Instruction}
#Input#
{Input}
\end{spverbatim}

\subsection{Code Generation Tasks}

\begin{spverbatim}
SYSTEM MESSAGE:
I want you act as a professional Prompt Rewriter.
USER PROMPTS:
Please increase the difficulty of the given programming test question a bit. You can increase the difficulty using, but not limited to, the following methods:
(1) Add new constraints and requirements to the original problem, adding approximately 10 additional words.
(2) Replace a commonly used requirement in the programming task with a less common and more specific one.
(3) If the original problem can be solved with only a few logical steps, please add more reasoning steps.
(4) Provide a piece of erroneous code as a reference to increase misdirection.
(5) Propose higher time or space complexity requirements, but please refrain from doing so frequently.

#Instruction#
{Instruction}
#Input#
{Input}
\end{spverbatim}

\section{GPT-4 Score Template}

\label{sec:score-template}

\begin{spverbatim}
SYSTEM MESSAGE:
We would like to request your feedback on the performance of an AI assistant. The assistant provides outputs for instruction and input (if any).
USER PROMPTS:
Please score the response to the instruction and input according to the following criteria.
The maximum score is 100 points, and it consists of 4 parts:
1. Clarity (15 points): Assign a score based on how effectively the instruction conveys the problem. High-quality, clear questions score higher.
2. Difficulty (25 points): Rate the complexity of the instruction's problem. Higher difficulty should receive a higher score.
3. Explanations (25 points): Assess if the response includes detailed explanations alongside any code provided. The more comprehensive the explanation, the higher the score.
4. Accuracy (35 points): Score the response based on the accuracy and correctness of the solution to the instruction's problem. Higher accuracy should receive a higher score.
Here's some examples and socres you can follow:
### Example 1: 
### Instruction: {EXAMPLE INSTRUCTION 1}
### Response: {EXAMPLE OUTPUT 1}
### Score for Example 1: {SCORE 1}
### Example 2:
### Instruction: {EXAMPLE INSTRUCTION 2}
### Input: {EXAMPLE INPUT 2}
### Response: {EXAMPLE OUTPUT 2}
### Score for Example 2: {SCORE 2}
### Example 3:
### Instruction: {EXAMPLE INSTRUCTION 3}
### Response: {EXAMPLE OUTPUT 3}
### Score for Example 3: {SCORE 3}

Please score the upcoming Instruction, Input and Response based on these examples across four dimensions, and then add the four scores together to get the total score. Try to avoid getting a full score as much as possible.
Please first output a single line containing the total score number only. 
In the subsequent line, please provide a comprehensive explanation of your evaluation, avoiding any potential bias.
### Instruction:
{INSTRUCTION}
### Input:
{INPUT}
### Response:
{OUTPUT}
\end{spverbatim}

\section{Quality Score Distribution}

\label{sec:quality-distribution}

\begin{figure}[ht]
  \centering
  \begin{subfigure}[b]{\linewidth}
    \includegraphics[width=\textwidth]{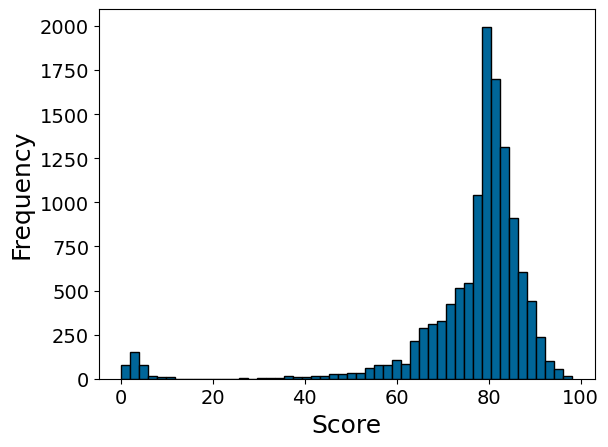}
    \caption{Quality Score Distribution in Code Generation Tasks (12k Data)}
    \label{fig:quality_subfig1}
  \end{subfigure}
  \hfill
  \begin{subfigure}[b]{\linewidth}
    \includegraphics[width=\textwidth]{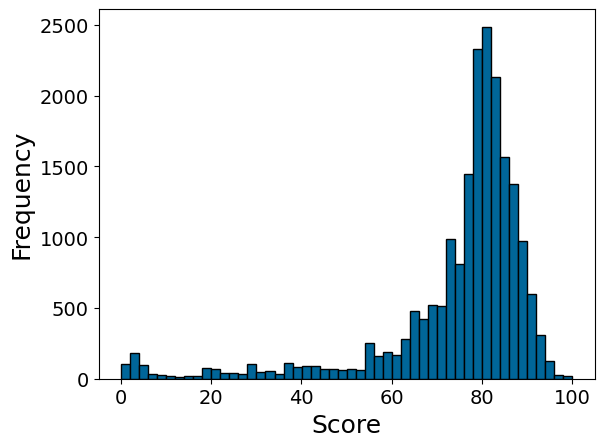}
    \caption{Quality Score Distribution in NLU Tasks (20k Data)}
    \label{fig:quality_subfig2}
  \end{subfigure}
  \caption{Quality Score Distribution}
  \label{fig:quality_scores_distribution}
\end{figure}

We have gathered the quality scores for the variety compressed dataset following the second step of our paradigm, both in code generation tasks and NLU tasks. The score distributions are depicted in Fig. \ref{fig:quality_scores_distribution}. Notably, the quality scores exhibit an approximately normal distribution within the score interval of 60 to 100 for both tasks. This observation validates the effectiveness of our scoring strategies in discerning low-quality data. It should be noted that the minor bumps near 0 stem from connection errors or OpenAI API calling ratio constraints, resulting in GPT scores of 0 for certain instructions.

\section{Experiments Implementation Details}

\label{sec:implementation_details}

For both foundation models, we conduct training on Azure Machine Learning Studio's cluster, utilizing 4 nodes, each equipped with 8 V100 GPUs featuring DeepSpeed Zero-3 \citep{rajbhandari-2019-zero} offload. Specifically, during the fine-tuning of Mistral 7B, we employ LoRA \citep{hu-2021-lora}. This strategy is chosen for its ability to ensure a more steady convergence of loss, resulting in better performance. The detailed fine-tuning arguments are outlined in Tab. \ref{tab:training_arguments}.

\begin{table}[ht]
    \caption{Fine-tuning Arguments for StarCoder 15B and Mistral 7B}
    \centering
    \begin{tabular}{lcc}
        \textbf{Arguments} & \textbf{StarCoder} & \textbf{Mistral}\\
        \hline
         model\_max\_length & 1024 & 2048 \\
         batch\_size & 8 & 8 \\
         num\_epoch & 3 & 3 \\
         learning\_rate & 2e-5 & 2e-5 \\
         fp16 & True & True \\
         lora\_r & - & 16 \\
         lora\_alpha & - & 32 \\
         lora\_dropout & - & 0.05 \\
         \hline
    \end{tabular}
    \label{tab:training_arguments}
\end{table}

\section{Benchmarks and Compared LLMs}

\subsection{Benchmarks}

\label{sec:benchmarks}

Below is an introduction to these chosen benchmarks:

\begin{itemize}
    \item \textbf{HumanEval} \citep{chen-2021-evaluating}. HumanEval is utilized to gauge functional correctness in synthesizing programs from docstrings. Comprising 164 original programming problems, it assesses language comprehension, algorithms, and simple mathematics.
    \item \textbf{MBPP} \citep{austin-2021-program}. The MBPP (Mostly Basic Python Problems) dataset consists of around 1,000 crowd-sourced Python programming problems. These are designed to be solvable by entry-level programmers, covering programming fundamentals and standard library functionality. In our experiments, to align with others, we select 400 questions.
    \item \textbf{HellaSwag} \citep{zellers-2019-hellaswag}. HellaSwag is a challenge dataset containing 70k multiple-choice questions for evaluating commonsense Natural Language Inference (NLI). While its questions may be trivial for humans (>95\% accuracy), they pose a challenge for state-of-the-art models.
    \item \textbf{ARC Challenge} \citep{clark-2018-think}. The AI2’s Reasoning Challenge (ARC) dataset is a multiple-choice question-answering dataset containing questions from science exams ranging from grade 3 to grade 9. It is split into two partitions: Easy and Challenge. The Challenge partition consists of 25k questions that require reasoning. 
    \item \textbf{TruthfulQA} \citep{lin-etal-2022-truthfulqa}. TruthfulQA is a benchmark designed to measure whether a language model is truthful in generating answers to questions. The benchmark comprises 817 questions spanning 38 categories. Questions are crafted so that some humans might answer falsely due to false beliefs or misconceptions.
    \item \textbf{MMLU} \citep{hendrycks-2020-measuring}. MMLU (Massive Multitask Language Understanding) is a new benchmark intended to measure knowledge acquired during pretraining. It evaluates models exclusively in zero-shot and few-shot settings, making it more challenging and akin to human evaluation. The benchmark covers 57 subjects across STEM, humanities, social sciences, and more, ranging in difficulty from elementary to advanced professional levels, testing both world knowledge and problem-solving ability.
\end{itemize}

\subsection{Compared LLMs}

\label{sec:compared llms}

The selected models for comparison in code generation tasks include:

\begin{itemize}
    \item \textbf{CodeT5+ \& InstructionCodeT5+} \citep{wang-2023-codet5+}. CodeT5+ is a new family of open code LLMs with an encoder-decoder architecture trained on various pretraining tasks. InstructionCodeT5+ is further fine-tuned on the Code Alpaca dataset.
    \item \textbf{Code LLaMA} \citep{rozire-2023-codellama}. Code Llama is a code-specialized version of Llama2 \citep{touvron-2023-llama2} trained on code-specific datasets.
    \item \textbf{StarCoder} \citep{li-2023-starcoder}. StarCoder is a widely-used large code language model trained on diverse sources, including 80+ programming languages, Git commits, GitHub issues, and Jupyter notebooks. It's also one of the foundation models in our paradigm experiments.
    \item \textbf{WizardCoder} \citep{luo-2023-wizardcoder}. WizardCoder is instruction fine-tuned on StarCoder with 78k instruction data generated through the application of Code Evol-Instruct.
\end{itemize}

The selected models for comparison in NLU tasks include:

\begin{itemize}
    \item \textbf{LLaMA} \citep{touvron-2023-llama}. LLaMA is a collection of foundation language models trained on trillions of tokens from publicly available datasets.
    \item \textbf{LLaMA2} \citep{touvron-2023-llama2}. Llama 2 is an updated version of Llama, trained on a new mix of publicly available data. It increased the size of the pretraining corpus by 40\%, doubled the context length of the model, and adopted grouped-query attention in training.
    \item \textbf{Mistral} \cite{jiang-2023-mistral}. Mistral is a state-of-the-art 7B foundational model, fast-deployed, easily customizable, and supports English and code with an 8k context length. It's also one of the foundation models in our paradigm experiments.
    \item \textbf{Vicuna} \cite{chiang-2023-vicuna}. Vicuna is an open-source chatbot trained by fine-tuning LLaMA on 70K user-shared conversations collected from the ShareGPT website.
    \item \textbf{WizardLM} \cite{xu-2023-wizardlm}. WizardLM is instruction fine-tuned on LLaMA with 70k instruction data generated through the Evol-Instruct strategy.
    \item \textbf{Platypus} \citep{lee-2023-platypus}. Platypus is a family of fine-tuned and merged LLMs achieving strong performance. It uses Open Platypus as its instruction dataset and applies the LoRA strategy to train adaptors that can be merged into different foundation models, creating many variant models.
\end{itemize}

\end{document}